 \documentclass[final,5p,times,twocolumn,authoryear]{elsarticle}

\usepackage{amssymb}
\usepackage{lipsum}

\journal{Elsevier}
\PassOptionsToPackage{hyphens}{url}

\usepackage{hyperref}
\usepackage{breakurl}
\usepackage{subcaption}

\begin{document}
\begin{center}
{\small
This is a preprint of a chapter accepted for publication in 
\textit{Challenges and Applications of Generative Large Language Models} (Pillai, A.S., Tedesco, R., Scotti, V., 2026. Challenges and Applications of Generative Large Language Models. Elsevier. ISBN: 9780443335921.).
The final published version will be available from Elsevier.
This preprint is posted with the explicit permission of the publisher.
}
\end{center}

\vspace{1em}

\begin{frontmatter}



\title{LLMs in Interpreting Legal Documents}


\author[first]{Simone Corbo}
\affiliation[first]{organization={Politecnico di Milano},
            city={Milano},
            country={Italy}}

\begin{abstract}
This chapter explores the application of Large Language Models in the legal domain, showcasing their potential to optimise and augment traditional legal tasks by analysing possible use cases, such as assisting in interpreting statutes, contracts, and case law, enhancing clarity in legal summarisation, contract negotiation, and information retrieval. 

There are several challenges that can arise from the application of such technologies, such as algorithmic monoculture, hallucinations, and compliance with existing regulations, including the EU’s AI Act and recent U.S. initiatives, alongside the emerging approaches in China. Furthermore, two different benchmarks are presented.
\end{abstract}




\end{frontmatter}




\section{Introduction}
\label{introduction}
The legal world is an area in which precision meets interpretation, and today, that delicate balance is being influenced by new technologies. As the complexity of legal texts grows and the demand for rapid, accurate analysis intensifies, Large Language Models (LLMs) have emerged as possible allies in streamlining manual legal tasks that can be extremely time-consuming. These models, capable of understanding and generating human-like text, can influence how we interpret statutes, contracts, legal documents and, most importantly, how the legal world is perceived by society.

To provide an example of the potential of this type of Artificial Intelligence, let us consider the American Bar Exam, a rigorous professional licensure test designed to assess legal knowledge and skills, it is the first step of the career of any American lawyer and it is designed to ensure that candidates master the legal principles necessary to practice law. This exam is known for its difficulty, it most commonly spans two days, with one day dedicated to the Multistate Bar Examination (MBE), which is a standardised 200-question test covering six key areas: Constitutional Law, Contracts, Criminal Law, Evidence, Real Property, and Torts \footnote{\url{https://www.americanbar.org/groups/legal_education/resources/bar-admissions/bar-exams/}.}

To qualify for the exam, candidates must complete at least seven years of post-secondary education, earning a four-year undergraduate degree followed by graduation from an American Bar Association-accredited law school. In addition to formal education, most applicants invest significant time and money in specialised test-preparation courses. Despite this effort, approximately 20\% of test-takers fail on their first attempt.

Following a study by \cite{katz_gpt-4_2023}, it has been shown that a zero-shot LLM (See  for a detailed explanation on zero shot-learning), specifically GPT-4, is capable of achieving a passing score on the Bar Exam. Moreover, test results indicate that its average performance exceeds that of human test-takers. Such a result is impressive to say the least and raises many questions and ethical concerns, mostly due to the impact that the judiciary system can have on one's life.

In this chapter, we are going to explore how Large Language Models can be used to aid legal-related tasks. To provide an introductory context we will explore the current legislation regulating its application, in particular we will mention the AI Act, the regulation used in the EU and the risks of an algorithmic monoculture. Then a series of case studies based on current publications are presented, showing four different use cases regarding the interpretation of laws using LLMs; the retrieval of information to clarify a vague concept, the negotiation of contracts and the use in creating legal summaries. Finally we will analyse two different benchmarks used in the evaluation of LLMs in legal tasks.
\section{Background}
To comprehend the size of the phenomenon we are dealing with, it can be helpful to look at some statistics, published yearly by the AI Index and Institute for Human-Centered AI by \cite{maslej2024}.
Considering the 2024 report we can have a complete overview on how this type of technology influences everyday aspects of our lives, providing data for topics such as Responsible AI, economy, science and medicine, education, policy and governance, diversity, and public opinion.
\subsection{Regulations}
As it is the very nature of the bodies of law to adapt to an ever changing society, lawmakers need to acknowledge the spread of such technology and its potential impact on vital aspects of the core entities that constitute the essence of a state and the judicial system is not an exception.

The three main driving forces of artificial intelligence, namely United States of America, European Union and China, have tackled this topic in different ways. In 2023, the United States experienced a sharp rise in AI regulations, marking a 56.3\% increase over the previous year and continuing a notable upward trend over the past five years and the most important national regulation was President Biden’s Executive Order on AI\footnote{\url{https://bidenwhitehouse.archives.gov/briefing-room/statements-releases/2023/10/30/fact-sheet-president-biden-issues-executive-order-on-safe-secure-and-trustworthy-artificial-intelligence/}}
The European Union adopted the AI Act in 2024, making the first example of such a wide regulation on the matter, the specific details will be presented in the next section.

As the EU chose a unified legal framework, the Chinese approach differs substantially, as it is considered agile\footnote{\url{https://www.holisticai.com/blog/china-ai-regulation}}, there are general industry-specific regulations and best practices built by AI governance pilot projects.

Chinese local authorities have issued experimental regulations but as for the current times, have not yet been adopted at the national level\footnote{\url{https://iapp.org/news/a/preparing-for-compliance-key-differences-between-eu-chinese-ai-regulations}}. 

\begin{figure}
     \centering
     \begin{subfigure}[b]{0.4\textwidth}
         \centering
         \includegraphics[width=\textwidth]{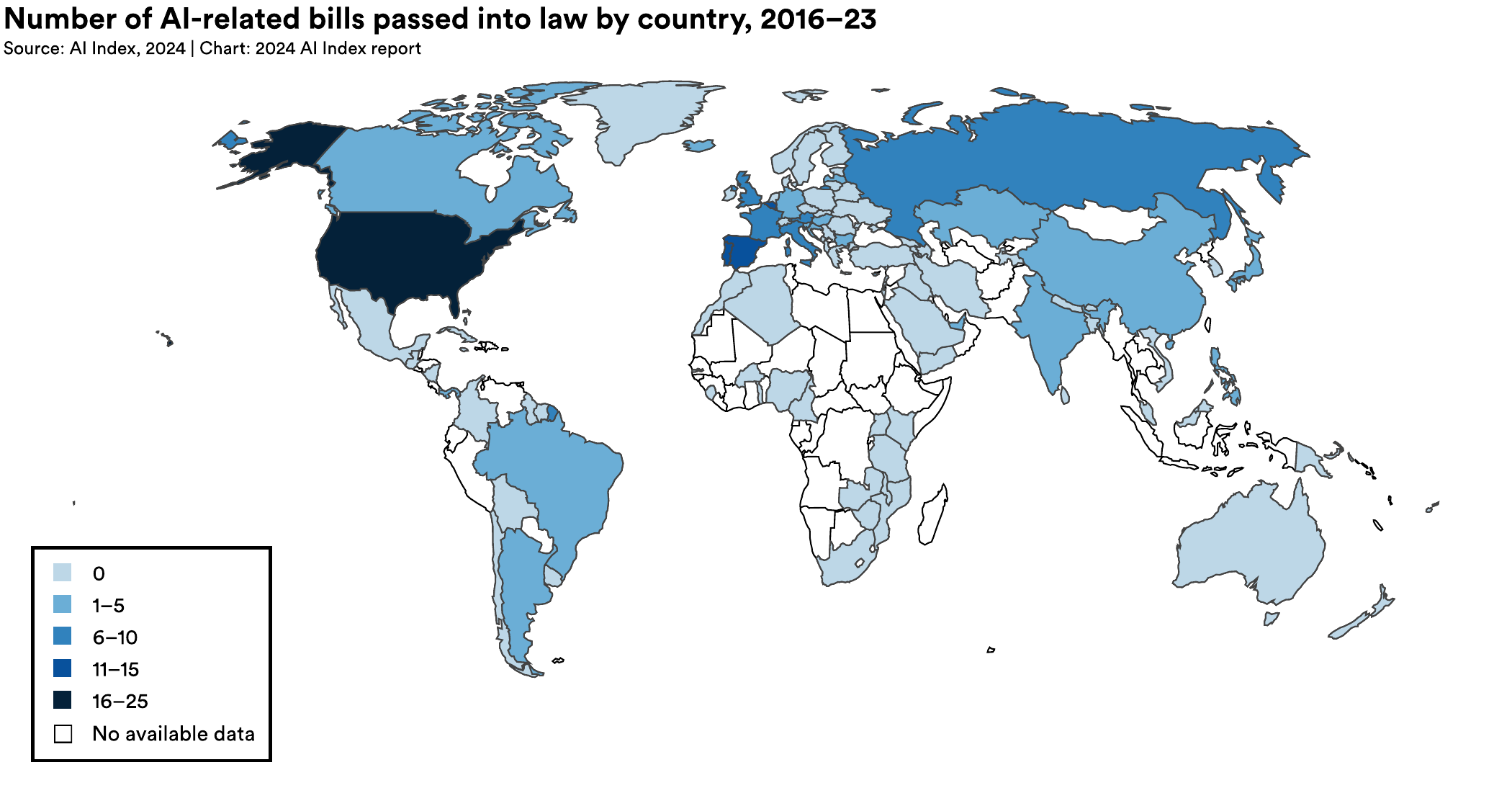}
         \caption{From Maslej, Nestor, Fattorini, Loredana, Perrault, Raymond, Parli, Vanessa, Reuel, Anka, Brynjolfsson, Erik, Etchemendy, John, Ligett, Katrina, Lyons, Terah, Manyika, James, Niebles, Juan Carlos, Shoham, Yoav, Wald, Russell \& Clark, Jack. (2024). The AI Index 2024 Annual Report. Licensed under Attribution-NoDerivatives 4.0 International.}
         \label{fig:mapBill}
     \end{subfigure}
     \hfill
     \begin{subfigure}[b]{0.4\textwidth}
         \centering
         \includegraphics[width=\textwidth]{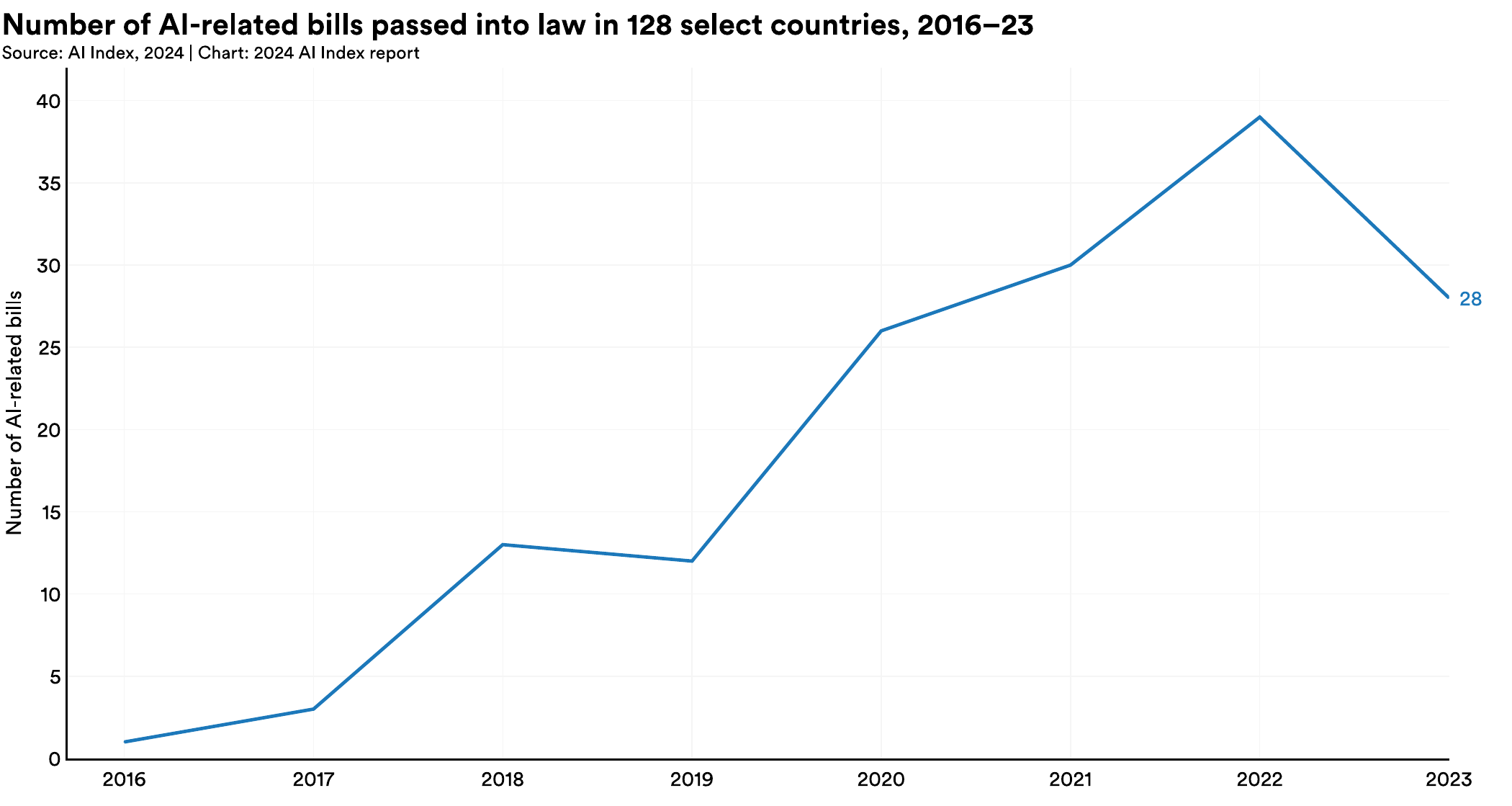}
         \caption{From Maslej, Nestor, Fattorini, Loredana, Perrault, Raymond, Parli, Vanessa, Reuel, Anka, Brynjolfsson, Erik, Etchemendy, John, Ligett, Katrina, Lyons, Terah, Manyika, James, Niebles, Juan Carlos, Shoham, Yoav, Wald, Russell \& Clark, Jack. (2024). The AI Index 2024 Annual Report. Licensed under Attribution-NoDerivatives 4.0 International.}
         \label{fig:numBill}
     \end{subfigure}
    \caption{Number of AI-related bills passed into law by country, 2016–23}
    \label{fig:bill}
\end{figure}

At the federal level, legislative attention soared: 181 AI-related bills were proposed in 2023, more than double the 88 proposed in 2022. Globally, the AI Index found legislative initiatives mentioning “artificial intelligence” in 128 countries from 2016 to 2023, with 32 nations enacting at least one AI-related bill (\ref{fig:bill}) for a total of 148 such bills. 

\begin{figure}
     \centering
     \begin{subfigure}[b]{0.4\textwidth}
         \centering
         \includegraphics[width=\textwidth]{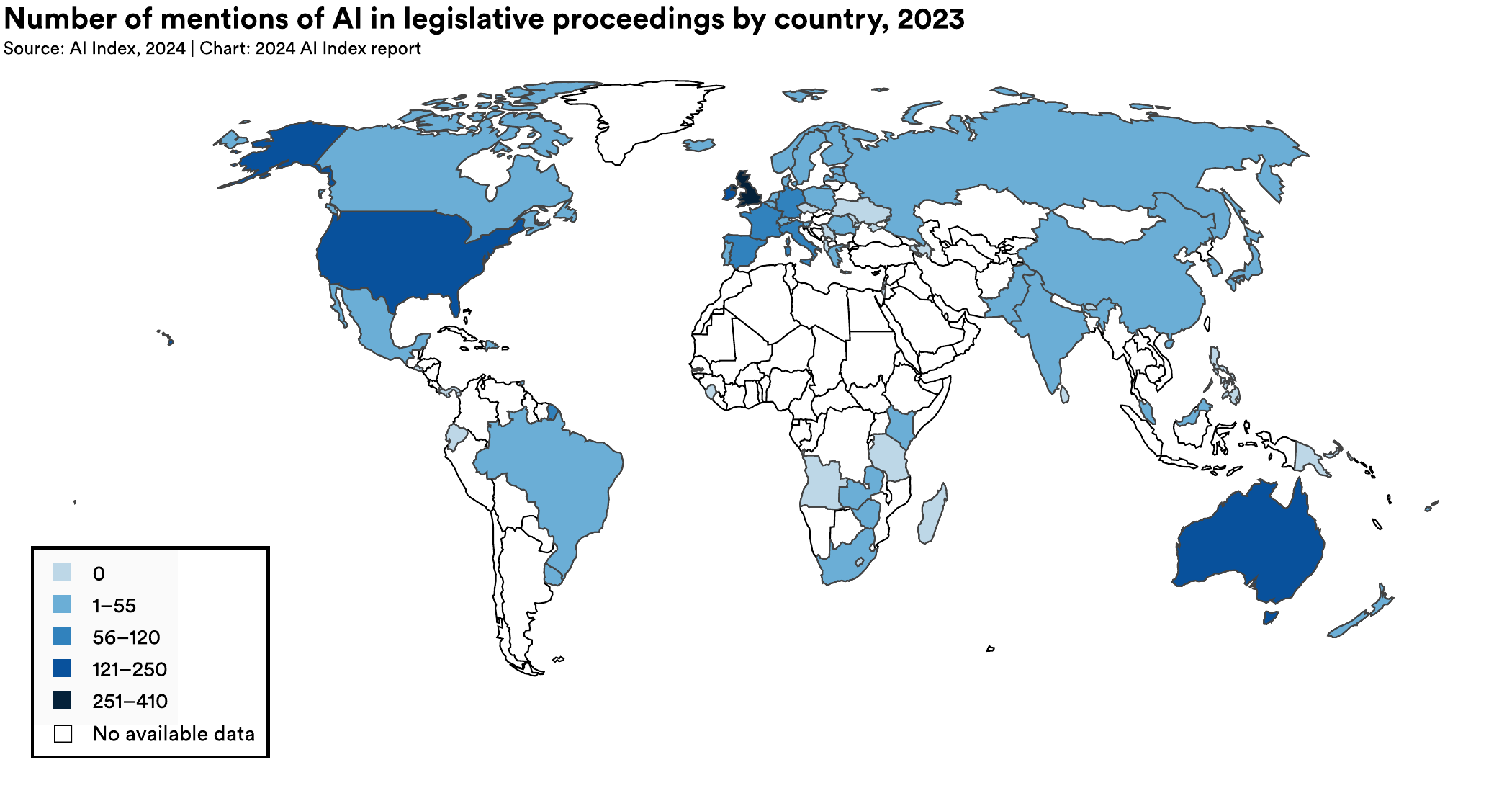}
         \caption{From Maslej, Nestor, Fattorini, Loredana, Perrault, Raymond, Parli, Vanessa, Reuel, Anka, Brynjolfsson, Erik, Etchemendy, John, Ligett, Katrina, Lyons, Terah, Manyika, James, Niebles, Juan Carlos, Shoham, Yoav, Wald, Russell \& Clark, Jack. (2024). The AI Index 2024 Annual Report. Licensed under Attribution-NoDerivatives 4.0 International.}
         \label{fig:mapLeg}
     \end{subfigure}
     \hfill
     \begin{subfigure}[b]{0.4\textwidth}
         \centering
         \includegraphics[width=\textwidth]{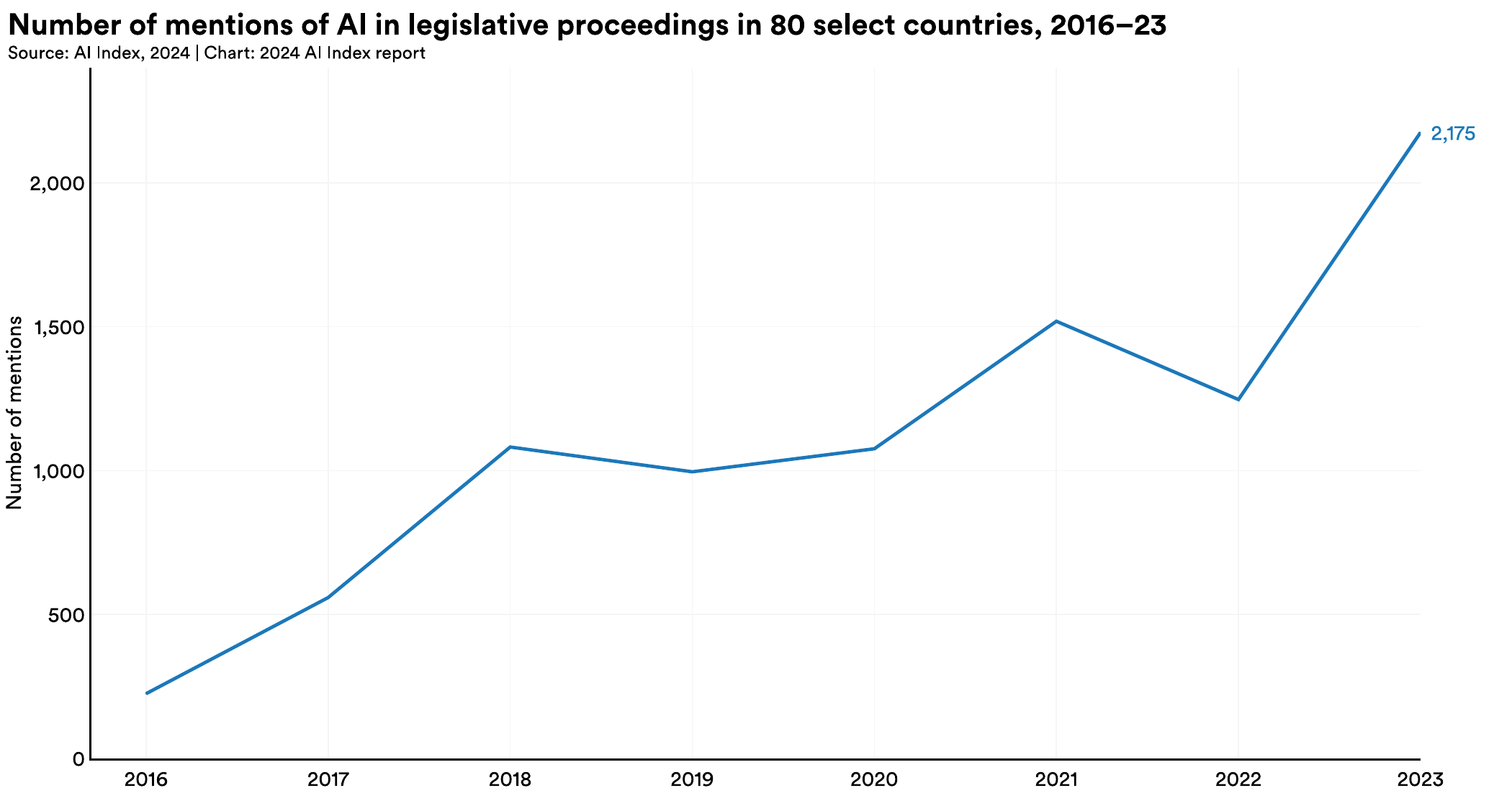}
         \caption{From Maslej, Nestor, Fattorini, Loredana, Perrault, Raymond, Parli, Vanessa, Reuel, Anka, Brynjolfsson, Erik, Etchemendy, John, Ligett, Katrina, Lyons, Terah, Manyika, James, Niebles, Juan Carlos, Shoham, Yoav, Wald, Russell \& Clark, Jack. (2024). The AI Index 2024 Annual Report. Licensed under Attribution-NoDerivatives 4.0 International.}
         \label{fig:numLeg}
     \end{subfigure}
    \caption{Number of mentions of AI in legislative proceedings}
    \label{fig:mentions}
\end{figure}

During the same period, mentions of AI in legislative proceedings nearly doubled, rising from 1,247 in 2022 to 2,175 in 2023 (\ref{fig:mentions}), spanning 49 countries and at least one from every continent.

\subsection{AI Act}
Large Language Models, such as ChatGPT, since its introduction to the public have been heavily scrutinised by its users, especially regarding sensitive topics that these systems might come across. One of these topics concerns its use in the legal field, while it can be helpful in many cases it has also been considered an “high-risk” AI system for the European Union, as per the definition provided by the AI ACT (Regulation 2024/1689), the regulation regarding AI systems operating within the EU.

More specifically, in Recital 61: “Certain AI systems intended for the administration of justice and democratic processes should be classified as high-risk, considering their potentially significant impact on democracy, the rule of law, individual freedoms as well as the right to an effective remedy and to a fair trial. In particular, to address the risks of potential biases, errors and opacity, it is appropriate to qualify as high-risk AI systems intended to be used by a judicial authority or on its behalf to assist judicial authorities in researching and interpreting facts and the law and in applying the law to a concrete set of facts. AI systems intended to be used by alternative dispute resolution bodies for those purposes should also be considered to be high-risk when the outcomes of the alternative dispute resolution proceedings produce legal effects for the parties. The use of AI tools can support the decision-making power of judges or judicial independence, but should not replace it: the final decision-making must remain a human-driven activity. The classification of AI systems as high-risk should not, however, extend to AI systems intended for purely ancillary administrative activities that do not affect the actual administration of justice in individual cases, such as anonymisation or pseudonymisation of judicial decisions, documents or data, communication between personnel, administrative tasks.”
\subsection{Algorithmic Monoculture}
Artificial intelligence is gaining more and more importance in legal decision-making and it raises the spectre of what has been defined as "algorithmic monoculture" in multiple fields as shown by \cite{kleinberg_algorithmic_2021}. In a legal setting, whether in risk assessments, case evaluations, or sentencing guidelines, this phenomenon emerges when multiple institutions rely on the same algorithmic framework. Even if this single algorithm is, taken individually, more accurate than its alternatives, a widespread adoption creates a uniform system vulnerable to correlated failures, much like the risks seen in biological monocultures.

This situation can be compared to the “Braess’ paradox” introduced by Arthur Pigou in 1920 (\cite{pigou_welfare_2017}) with reference to the flow of traffic on a road network. The choice of the preferred route depends not only on the quality of the road but also on the existing traffic. If drivers behave in an “egoistic” way by taking the path that is optimal for them, the general running times might not be minimal and oddly enough, the extension of a road, such as adding a new lane, may lead to a redistribution of the traffic that worsens the individual running times.

By making the comparison to decision-making algorithms, the paradox implies that a more accurate algorithm might improve the average quality of decisions when used by a single decision-maker, its universal application across a network of legal agents can inadvertently reduce overall decision quality. The cumulative effect is that the system, as a whole, becomes less robust, and even subtle, low-level errors can compound over time.

Under “normal” conditions, an individual legal decision-maker might benefit from the improved accuracy of a well-designed algorithm. However, when all legal bodies converge on this single solution, the risk profile shifts dramatically. Even without any external shocks or dramatic events to unmask the vulnerabilities the system may experience a decline in performance. 

This decline is not necessarily dramatic enough to be immediately noticeable, yet it can erode the quality of collective legal outcomes over time, especially by creating dangerous precedents.

Despite the fact that more accurate algorithms enhance decision-making in isolated scenarios, the lack of heterogeneity in these types of processes allows valuable options and different interpretations to slip through the cracks. In legal applications, this could mean overlooking crucial details in a case, missing extenuating circumstances, or failing to recognise emerging legal trends that do not fit within the standardised model. The result is a reduction in overall social welfare, as the benefits of diverse perspectives and approaches are lost.
\section{Legal Impact and Uses}
The versatility of LLMs for a variety of tasks is considered as one of the key aspects of its commercial success and the legal field is not an exception, quite the contrary, as in recent years numerous different applications have been explored.

\begin{figure}
    \centering
    \includegraphics[width=1\linewidth]{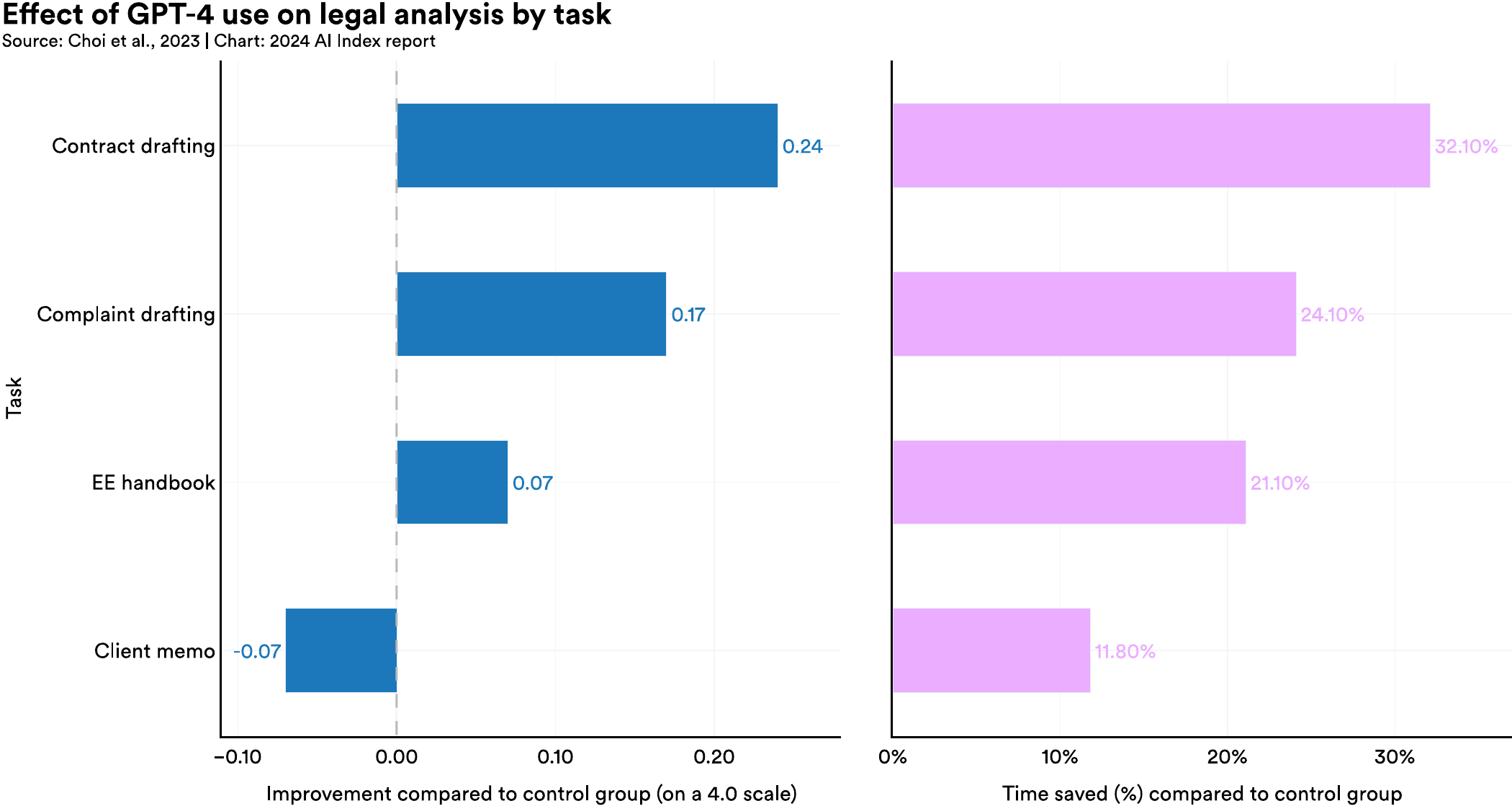}
    \caption{Comparison between two groups in various tasks, divided by Improvement and Time Saved. From Maslej, Nestor, Fattorini, Loredana, Perrault, Raymond, Parli, Vanessa, Reuel, Anka, Brynjolfsson, Erik, Etchemendy, John, Ligett, Katrina, Lyons, Terah, Manyika, James, Niebles, Juan Carlos, Shoham, Yoav, Wald, Russell \& Clark, Jack. (2024). The AI Index 2024 Annual Report. Licensed under Attribution-NoDerivatives 4.0 International.}
    \label{fig:effect}
\end{figure}

A study on the use of AI in a legal setting (\cite{maslej2024}) presented the results that teams with GPT-4 access significantly improved in efficiency and showed a dramatic improvement in various legal tasks, especially contract drafting. \ref{fig:effect}  illustrates the improvements observed in the group of law students who utilised GPT-4, compared to the control group, in terms of both work quality and time efficiency across a range of tasks. Although AI can assist with legal tasks, there are also widespread reports of LLM hallucinations being especially pervasive in legal tasks.

The possible uses of Generative AI in this field can be found in interpreting laws in order to define a specific term that is considered crucial in a specific article's field of applicability, to retrieve relevant information from a database, to testing contract clauses and to provide accessible summaries of court decisions.

In the next paragraphs a few use cases are reported, taken from different research articles that have subsequently found real world implementations and are currently used by law firms and providers around the world.
\subsection{Interpreting Laws}

The introduction of ChatGPT has gained much traction within the general public, becoming the 8th most visited website as of January 2025\footnote{According to \url{https://en.wikipedia.org/wiki/List_of_most-visited_websites}}, a telltale sign of the revolution that this technology holds is its versatility, as it can be used to retrieve information with high accuracy, rewrite paragraphs, effortlessly code complex applications in an instant and most importantly, its use has been shown to be useful even in a trial.

A significant example is its use in a court case\footnote{Snell v. United Specialty Insurance Company, 102 F.4th 1208 (11th Cir. 2024)  \url{https://media.ca11.uscourts.gov/opinions/pub/files/202212581.pdf}}, in which a Judge used ChatGPT to clarify the meaning of the word “landscaping” to help the decision regarding which tasks can be considered landscaping as it became the pivotal point in the case. Specifically he asked: “What is the ordinary meaning of  ‘landscaping’?” to which the model replied: “'Landscaping' refers to the process of altering the visible features of an area of land, typically a yard, garden or outdoor space, for aesthetic or practical purposes. This can include activities such as planting trees, shrubs, flowers, or grass, as well as installing paths, fences, water features, and other elements to enhance the appearance and functionality of the outdoor space.”

Following the discussion contained in the court transcripts, the use of ChatGPT was seen as an experiment and Judge Kevin C. Newsom expressed both positive and negative points to the use of LLMs in interpreting the meaning of specific words in a legal setting.
\subsubsection{Pros}

In a system of Common Law, like the one in the United States, the legal precedents are considered when handing out a sentence, and the clarity of the words used in writing such laws has been a crucial point in the research done by law scholars. “Words are to be understood in their ordinary, everyday meanings—unless the context indicates that they bear a technical sense.” \cite{scalia_reading_2012}. Due to the intrinsic nature of the training process of LLMs, they are trained on ordinary-language inputs, so it can be considered that the meaning of the words interpreted by the model is comparable to the one of ordinary use.

Judiciary cases where the precise definition of a word carries significant weight, such as the one previously mentioned, is not uncommon. In such situations, judges often refer to dictionaries to determine the most appropriate interpretation based on the context. The issue of transparency also extends to the definitions found in commercially available dictionaries, as they are crafted by humans, and the selection of one definition over another inherently involves a degree of discretion. Moreover, judges rarely provide a detailed explanation of their decision-making process, as noted in the reasoning: “they rarely explain in any detail the process by which they selected one definition over others. Contrast my M.O. in this case, which I would recommend as a best practice: full disclosure of both the queries put to the LLMs (imperfect as mine might have been) and the models’ answers".

The availability of tools such as ChatGPT is one of the key points of its success, along the fact that it is free which is opposed to some of the most famous online dictionaries that have a paywall.

Recently, the dictionary approach has been criticised as there has been some wide-ranging surveys of ordinary citizens, which showed that dictionaries do not always capture ordinary understanding of legal texts (\cite{tobia_testing_2020}). Others have turned to the study of corpus linguistics. “Corpus linguists study language through data derived from large bodies, corpora, of naturally occurring language. They look for patterns in meaning and usage in large databases of actual written language […] with the goal of quantifying the patterns  of words' usages and occurrences in large bodies of language" \cite{lee_judging_2017}. This last method has been criticised as those tasked with compiling the data might introduce bias in selecting the inputs (\cite{choi_measuring_2022}). As much as the survey method offers validation from the public, it is widely impractical, as judges and lawyers have neither the time nor the resources to poll ordinary citizens on a regular basis. By contrast, as said before, LLMs are both available and accessible.

Another point in favour of its use in this type of task, is that LLMs can “understand” context. Even with polysemic words, such as “bat”, the models can interpret which meaning needs to be considered.

\subsubsection{Cons}
LLMs can “hallucinate”. Hallucination is typically referred to as a phenomenon in which the generated content appears nonsensical or unfaithful to the provided source content (\cite{huang_survey_2023}). There has been a case in which a lawyer used ChatGPT to write a brief submitted in a trial but it was promptly rejected as it contained more than half a dozen relevant court precedents that didn't exist\footnote{Benjamin Weiser, (May 27, 2023), Here’s What Happens When Your Lawyer Uses ChatGPT, The New York Times, \url{https://www.nytimes.com/2023/05/27/nyregion/avianca-airline-lawsuit-chatgpt.html}}. With the progress of the technology regarding LLMs, the phenomena of hallucinations will become fewer and farther between. Hallucinations can have a more relevant impact when the question asked requires a specific answer, rather than the lookup of a dictionary entry. One last aspect to consider is that also human lawyers “hallucinate” in both good faith and intentionally. Moreover the hallucination issue advocates against blind-faith reliance on LLMs, in the same way as a judge would not rely blindly on a lawyer.

Another issue that can arise is that LLMs do not capture offline speech, and thus might not fully account for underrepresented populations’ usages. The texts used in the training of LLMs might not represent accurately all the minorities or people in rural areas that might not have Internet access and their written speech is less likely to be included in the LLMs’ ordinary-meaning assessment. Another fact to be noted is the “regionalised” meaning that certain words have acquired. For example according to the \cite{mw} dictionary, “toboggan” can refer to either “long flat-bottomed light sled,” “downward course or sharp decline,” or a “stocking cap.”. Noting that the third sense is a regional one, when the Judge asked ChatGPT, “What is the ordinary meaning of ‘toboggan’?”, it responded with only the first, sled-based explanation.

Lawyers, judges, and would-be litigants might try to manipulate LLMs: As LLMs would become relevant in the legal panorama, there is the risk of lawyers and judges to reverse-engineer a prompt that gives out a preferred answer, as it is already the case with the choice of advantageous dictionary definitions. A question that was raised by the Judge is whether prospective litigants might corrupt the training data on which the LLMs are trained on, in order to give out a preferred interpretation. In light of the fact that the same AI companies that develop this kind of technology might be the mentioned prospective litigants. Considering that most models that are available online have a "cutoff date", it would likely be difficult, if not impossible, to pollute the inputs retroactively. Another factor that might mitigate this risk is the enormous size of the training corpus, which makes an attack of this kind extremely difficult. To overcome this scenario, an idea could be to query different models and compare the responses.

A final consideration was noted by Judge Newsom and it concerns the importance of human judgment in interpreting results generated by LLMs, avoiding the complete reliance of the judiciary system to AI models as it can lead to dystopian scenarios.
\subsection{Retrieving Information}

After interacting with an LLM, especially online chatbots, most people get the impression that they are engaging with an all-knowing entity capable of answering every question we might have with incredible precision and accuracy. However, when we consider the basic structure of a language model and its probabilistic nature it should come as no surprise that delivering always a correct answer is not possible, especially with historical or factual data. Accurate responses are common, but users should keep in mind that precise references may reflect probabilistic guesses.

\subsubsection{Hallucinations and Legal Systems}

\begin{figure}
    \centering
    \includegraphics[width=1\linewidth]{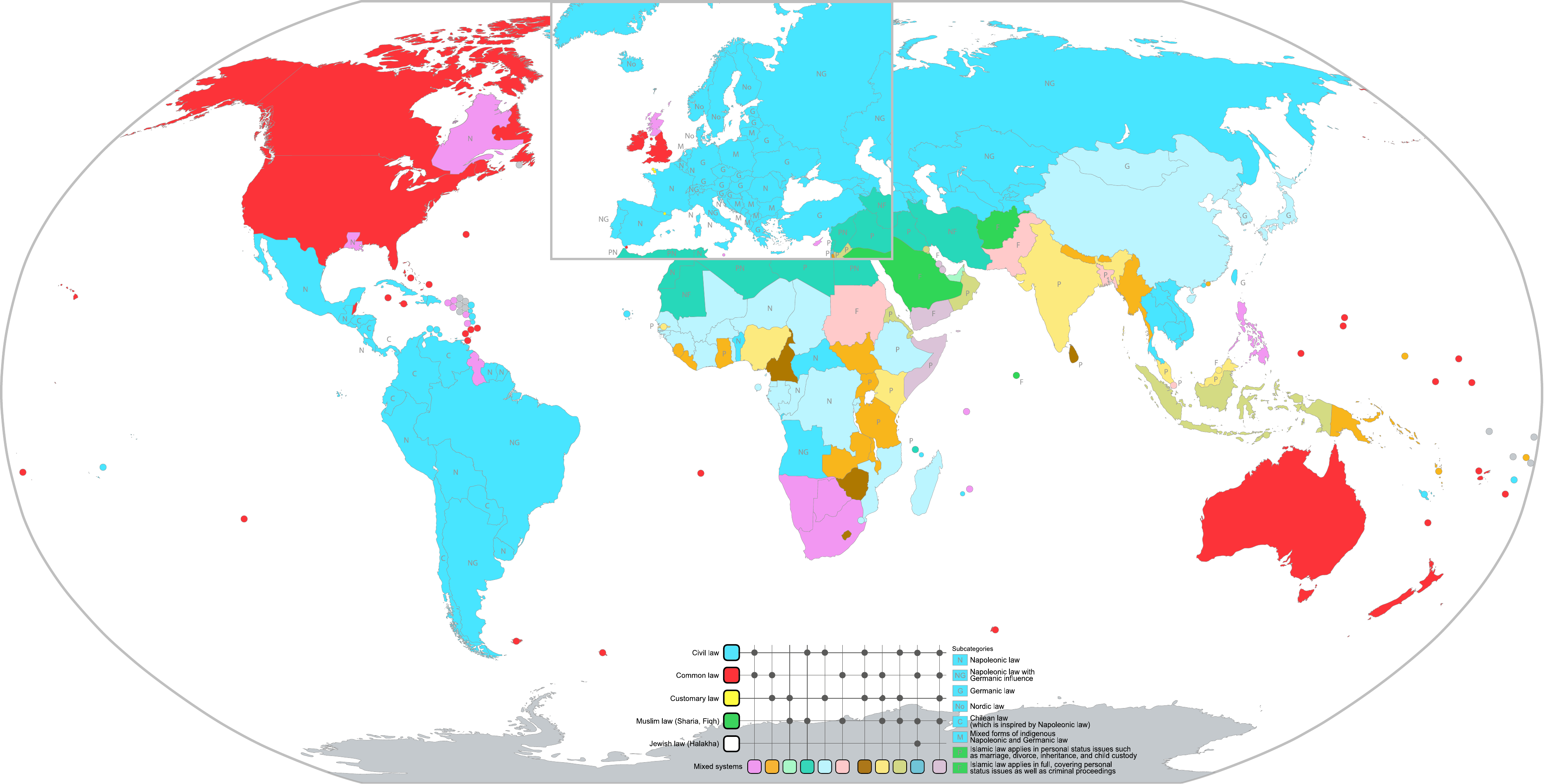}
    \caption{Civil law-based systems are in blue, Common law-based systems are in red and mixed systems incorporating elements of both civil law and common law in pink. Derivate: Goldsztern, Original: Maximilian Dörrbecker, Public domain, via Wikimedia Commons}
    \label{fig:systems}
\end{figure}

When interpreting laws and past cases to predict the outcome of a trial, a crucial factor is the legal system of the country in question. Among the various legal systems adopted in the world (see \ref{fig:systems}), the two most common are Common Law, in which judicial precedent established by previous cases and legal codes are considered in a ruling, and Civil Law, where legal codes serve as the primary source of law with less emphasis on precedent as they do not constitute binding legal value.

The concept of hallucination in LLMs becomes even more relevant in cases of Common Law systems, in which citing a non-existent precedent would render a legal document completely invalid. The concept of hallucination is a cardinal point in the reliability in LLMs, a more formal approach is defined in Chapter 7, in the next paragraphs a few notable examples relative to its impact on the legal domain are shown.

A study published by Stanford RegLab and Institute for Human-Centered AI (\cite{dahl_large_2024}) presents this phenomenon by testing 4 popular off-the-shelf models with three batches of tests of increasing difficulty, reported here:
\begin{itemize}
    \item Low: This type of tasks do not require higher-order legal reasoning but rather tests the internal knowledge and ability to recall information from the case syllabus. The question asked was: 
    \begin{itemize}
        \item Existence: Given the name and citation of a case, state whether the case actually exists or not;
        \item Court: Given the name and citation of a case, supply the name of the court that ruled on it;
        \item Citation: Given a case name, supply the Bluebook\footnote{The Bluebook: A Uniform System of Citation is a style guide that prescribes the most widely used legal citation system in the United States. (\url{https://en.wikipedia.org/wiki/Bluebook)}} citation of the case;
        \item Author: Given the name and citation of a case, supply the name of the opinion author. 
    \end{itemize}
    \item Moderate: Unlike the previous set, there is the requirement to evince knowledge of actual legal opinions themselves. Examples are such:\begin{itemize}
        \item Disposition: Given a case name and its citation, state whether the court affirmed or reversed the lower court;
        \item Quotation: Given a case name and its citation, supply any quotation from the opinion;
        \item Authority: Given a case name and its citation, supply a case that is cited in the opinion;
        \item Overruling year: Given a case name and its citation, supply the year that it was overruled.
    \end{itemize}
    \item High: High complexity tasks require  LLM to synthesise core legal information out of unstructured legal prose, it is frequently the topic of deeper legal research. An example of the tasks is the following:
    \begin{itemize}
        \item Doctrinal agreement: Given two case names and their citations, state whether they agree or disagree with each other;
        \item Factual background: Given a case name and its citation, supply its factual background;
        \item Procedural posture: Given a case name and its citation, supply its procedural posture\footnote{How did the case get to this court? What were the legal claims and what happened in the lower court (from \url{https://law.uc.edu/content/dam/refresh/law/pdfs-temporary/Orientation\%202018\%20casebrief\%20(1).pdf)}};
        \item Subsequent history: Given a case name and its citation, supply its subsequent procedural history, if any;
        \item Core legal question: Given a case name and its citation, supply the core legal question at issue;
        \item Central holding: Given a case name and its citation, supply its central holding.
    \end{itemize}
\end{itemize}

The final results presented in the benchmark, show that the results are dependent on a few factors, such as:
\begin{itemize}
    \item Task Complexity: as one could imagine the LLMs perform progressively worse as the difficulty increases, the best results are obtained on the simple Existence task. The models start to fail to provide information about a case’s Court, Citation, or Author. The worst result was achieved on Doctrinal agreement, the LLMs’ hallucination rates on this task, near 0.5, represent little improvement over random guessing;
    \item Court: By analysing the Court level from which the various cases appear from (starting from USDC, the lowest and more local level, up to the Supreme Court, the highest), LLMs do not perform well on localised legal knowledge. After all, the vast majority of cases are not appealed up to the Supreme Court, thus making them less likely to be exposed the material used in the training phase of LLMs;
    \item Jurisdiction: The United States Judicial System is divided into 13 jurisdictions and can hear both civil and criminal cases. The best performance was obtained when cases from the Ninth Circuit (comprising California and adjacent states), the Second Circuit (comprising New York and adjacent states), the Third Circuit (comprising Pennsylvania and adjacent states), and the First Circuit (comprising Maine and adjacent states) and performance tends to  degrade in circuits in the centre of the country. These results are influenced by the historic role that the Second, Third, and Ninth Circuits play in the American legal system and the relevance that they represent on a national level. Surprisingly, tests regarding cases from the D.C. Circuit, which is generally thought to be the most influential appellate division, obtained bad results. This counterintuitive finding is one example of the way that unanticipated biases might trouble the reliance on LLMs in practice;
    \item Case Prominence: Case prominence is negatively correlated with hallucination;
    \item Case Year: The examined models best results are found on cases that are several years behind the current state of the doctrine, and LLMs may fail to correctly deal with very old cases, but with still applicable and relevant laws involved;
    \item LLM: Not all models perform in the same way, in the benchmark tests considered, GPT 4 performs best overall, followed by GPT 3.5, followed by PaLM 2, followed by Llama 2.
\end{itemize}

After examining the possible factors that might affect the accuracy of the information retrieved, there are two other possible biases that were noted by the authors: Model sycophancy, defined as “The phenomenon where a model seeks human approval in unwanted ways” (\cite{sharma_towards_2023}) and General Cognitive Error, as when a legal researcher poses a question, it might introduce a bias as the question itself might contain factually incorrect references, maybe based on folk wisdom regarding some legal aspects. This phenomenon can be observed in pro se litigants that do not possess enough legal knowledge to be aware that the prompt contains non-factual information. Subsequently, there is a risk that the model accepts the users' misconception as legally based and produce related hallucinations in return.
\subsubsection{RAG}

There is a way to incorporate precise references or information from a database, this method is called Retrieval-Augmented Generation and it was introduced in 2020 by Meta (\cite{lewis_retrieval-augmented_2020})
As Luis Lastras, director of language technologies at IBM Research, said: “It’s the difference between an open-book and a closed-book exam, […] In a RAG system, you are asking the model to respond to a question by browsing through the content in a book, as opposed to trying to remember facts from memory.”\footnote{\url{https://research.ibm.com/blog/retrieval-augmented-generation-RAG}}

A basic implementation of RAG is composed by three operations:
\begin{enumerate}
    \item Indexing: Raw data is extracted from different formats and converted into plain text and it is divided into smaller chunks. With an embedding model, chunks are then encoded into their vector representation and stored into a database (e.g. using an SQL-based database, the following information is saved: (text, source, embedding)). This operation is done asynchronously with respect to receiving a query from the user.
    \item Retrieval: This operation, unlike the previous, is done when a query is received and using the same embedding model as the one used in the indexing phase, the query is transformed into its vector representation. Then for each entry in the indexed corpus and the vectorised query, the similarity score is computed and the top K chunks are selected and used to expand the context in the prompt.
    \item Generation: The query, with the addition of the newly found context, is created and is given as input to the model
\end{enumerate}

Since its first introduction, this approach has been expanded, such as introducing the Advanced RAG and Modular RAG by \cite{gao_retrieval-augmented_2023}
\subsubsection{Legal Example: Vague Concepts in Laws}

When dealing with legal terms and laws, the exact wording might leave some discretion and unclearness to the eyes of the reader, especially whom are not familiar with the technical vocabulary.

A possible use of RAG towards a more clear comprehension was proposed by \cite{luo_automating_2025}, rather than defining a specific term using general knowledge, as seen in the first use case, it employs a Retrieval-Augmented Generation to retrieve cases that are relevant to the vague concept from a case database, and then extracts concept-relevant key information from these cases to generate the concept interpretation and summarise the extracted information.

An example can be found in the Criminal Law of the People’s Republic of China, the article 264 corresponding to the crime of Theft, which states: “Whoever [...] enter a dwelling to steal [...], shall be sentenced to imprisonment of not more than 3 years, [...]”. The term "dwelling" can be considered a vague concept, as the article does not provide a clear definition of what kind of places are considered as such.

To provide an answer to the query the process can be split into three phases:
\begin{enumerate}
    \item Retrieve: In this phase the main goal is to find case judgments that mention the vague concept. Formally, given a vague concept $c$ and the article $a$ that the concept belongs to, a search of the database that stores previous judgments is performed in order to find all the cases judgments that contain the article number $a$. Then all the cases that mention the concept are retrieved through exact string matching and all of the retrieved cases form the set called $\mathcal D_0$. In RAG there could be different approach to the retrieval stage, but in this case the exact match was considered sufficient as legal terminology in the dataset is rigorous.
    \item Filter and Extract: In this stage there is the definition of whether a case is relevant, based on the presence of a detailed reason why the vague concept is applied to the considered case. The process of filtering is important, as not all judgments that mention a concept can be used to generate its interpretation. The LLM takes as input the court view, to determine whether it provides a detailed reason  for which the concept  applies. If said reason exists, then the LLM is prompted to determine, via a binary label $\mathcal l$ (Yes/No), whether the concept applies to the initial case. Therefore a refined set of $\mathcal D_0$ is obtained, called $\mathcal D_1$. The researchers noted that the number of positive cases was on average 100 times the number of negative cases (where does not apply) due to a series of factors related to judicial practices, therefore to guarantee equal attention to both cases, a balanced subset $\mathcal R$ of positive and negative examples is sampled.
    \item Interpret: The LLM is used in this phase to summarise all the retrieved information. The input given to the LLM is made of: legal article $a$, vague concept $c$, reason set $\mathcal R$, interpretation example $\mathcal e_0$ (composed of a vague concept $c_0$, corresponding article $a_0$ and reason set $\mathcal R_0$ and an interpretation selected apriori by a legal expert). The interpretation is composed by three main components:
    \begin{enumerate}
        \item Analysis: Basic meaning of the concept and its applicability conditions;
        \item Case Examples: Representative positive and negative cases from past rulings;
        \item Judicial Discretion: Criteria to guide judges in flexibly applying vague concepts based on case specifics.
    \end{enumerate}
\end{enumerate}
\subsection{Contract Negotiations}

Contracts are often derived from standardised templates that are used as the base for negotiations and legal agreements. However, real-world contracts frequently deviate from these templates to embrace specific circumstances or negotiated changes. Such deviations can carry significant legal and commercial implications and often it is hard for legal professionals to accurately identify and understand these differences.

A use case that might be relevant is the one presented by \cite{narendra_enhancing_2024} which is designed to automatically compare contracts with their corresponding templates, identifying where they align and where they diverge. By breaking down templates into key concepts and testing contract clauses for alignment, or amendments, this implementation offers a systematic approach to highlighting critical differences.

The introduction of such a tool is crucial for efficiency and accuracy in legal document analysis, as this task is considered extremely time-consuming and requires extended knowledge in the field of contract law. It also provides a foundation for refining standard templates but ultimately, this automated comparison approach supports legal professionals in managing complex documents and ensuring that every deviation is clearly understood and appropriately addressed. 

In order to complete the task, two components are necessary:
\begin{itemize}
    \item Natural Language Inference: first splits a template into key concepts and then uses LLMs to decide if the concepts are entailed by the contract document. The process can also be repeated in the opposite direction: contract clauses are tested for entailment against the template clause to see if they contain additional information. The first task involves classifying the relationship between a contract and a set of hypotheses. Each hypothesis is a single sentence, and the goal is to determine whether the hypothesis is entailed by, contradicts, or is neutral with respect to the contract.
    \item Evidence Extraction: The second task focuses on identifying evidence within the contract that supports the classification decision made in the first task.
\end{itemize}

By the nature of the first task, the non-entailed concepts are stored into a clause library, therefore allowing the building of  a catalog of approved contract terms based on historical contracts. A Retrieval Augmented Generation (RAG) pipeline is able to provide helpful references in contract management, particularly in reviewing contracts against a template to compare clause variations.

One of the primary challenges is to identify the variabilities of specific clauses in the master contract agreements compared to the template master agreements. The clauses analysed include:
\begin{itemize}
    \item Limitations of Liability: This is a clause that caps the amount of damages one party can claim from the other in case of a breach or other legal issue.\footnote{\url{https://www.icertis.com/contracting-basics/limitation-of-liability-clause}}
    \item Insurance: Contractual clause that requires one or both parties to maintain specific insurance coverage to mitigate risks associated with the contract.\footnote{\url{https://www.spellbook.legal/clauses/insurance}}
    \item Indemnity: Clause that specifies that one party (the indemnifying party) will compensate the other party (the indemnified party) for any losses or damages that may arise from a particular event or circumstance.\footnote{\url{https://ironcladapp.com/journal/contracts/indemnification-clause/}}
    \item Representations and Warranties: Clause that contains factual statements and assurances about conditions relevant to the contract. Representations address current or past facts, while Warranties promise the ongoing truth or legal backing of these facts\footnote{\url{https://www.spellbook.legal/clauses/representations-and-warranties}}
    \item Red Flags: These are provisions that create unbalanced obligations, sow confusion, or leave your company vulnerable in unforeseen circumstances\footnote{\url{https://www.contractzy.io/blog/red-flag-clauses-what-legal-counsel-needs-to-look-for-in-contracts}}
    \item System Modifications: Contractual clause that specifies the process and conditions for amending a contract’s terms\footnote{\url{https://www.spellbook.legal/clauses/modification}}
    \item Assignment: This clause gives a party the chance to engage in a transfer of ownership or assign their contractual obligations and rights to a different contracting party.\footnote{\url{https://www.contractscounsel.com/g/167/us/assignment-clause}}
    \item Source Code Escrow: An arrangement between the licensor and a licensee of software in which the licensor deposits a copy of the software's source code (and related technical components and documentation) with an independent escrow agent.\footnote{\url{https://uk.practicallaw.thomsonreuters.com/6-502-4093?transitionType=Default&contextData=(sc.Default)&firstPage=true}}
    \item Audits: Grants one party the right to examine and verify the other party's financial records, ensuring compliance with contractual terms.\footnote{\url{https://www.contractken.com/glossary/audit-clause}}
\end{itemize}

By comparing these clauses between the provided agreement and the template agreements, is becomes possible to understand the common deviations and variations that occur during contract negotiations and amendments.

Amendments are vital in the evolution and maintenance of legal contracts. They offer a formalised way to update contracts as there is a change in circumstances, ensuring that the terms remain valid and enforceable over time. By incorporating amendments, the parties involved can address unforeseen issues, clarify ambiguous language, or adjust responsibilities without having to sign a new contract. However, this adds another challenge to use LLMs to modify master contracts by incorporating amendments.

One key observation was that the best performing model is GPT-4, as it requires advanced text processing ability. To address this, a specific chunking of the document using a fine-tuned model is used, which helped in breaking down the document into various subsections. The process involved the following stpdf:
\begin{enumerate}
    \item Summarising Amendments: The first step is to create a summary of the amendment document.
    \item Extracting Key Data: Upon extracting the relevant sections and associated text from the amendments in JSON format, the modified master contract, incorporating these amendments, was generated. 
    \item Concept Extraction from Template Clauses: To further analyse the clauses, the template master agreements were divided into multiple concepts or hypotheses. This step allowed us to break down each template clause into its fundamental concepts, making it easier to compare and analyse against the master agreements. The term "concept" refers to a specific segment of the original clause, maintaining the integrity and context of the clause. Each clause is divided into multiple concepts.
\end{enumerate}

Once the concepts are extracted from each template clause, the chunks embedded and saved, a RAG pipeline can be developed to ask question of the document for each concept in template clause. For each chunk retrieved in response to the above question, cross-references to other sections are appended to the chunk. This approach ensured that it is possible to accurately determine whether each concept was present in the contract document, providing a comprehensive analysis of clause coverage and variability.
\subsection{Legal Summarisers}

As we have seen before, often the language used in legal settings can be difficult to interpret. Such jargon is often times called “Legalese” as it constitutes a parallel language with respect to plain English, with its specific vocabulary and phrases.

Despite the need for precise and technical language, it is not accessible to non-legal readers and makes the general public lose trust in the judicial system, therefore the need of summaries and explanations of high-profile cases has became a central priority in the field of legal communication.

The process of writing such texts is a long task and requires the work of experienced lawyers, making this effort worth only for the most important cases, but the next use case we are going to explore is this exact research topic as proposed by \cite{ash_translating_2024}. The results show that it is possible to make complex judicial texts accessible even to readers with relatively low levels of education.

Before commencing the explanation of the specific use case, it is worth explaining what is the task we are studying. Automated summarisation can be done via two different methods, Extractive summarisation, composing the summary by extracting keywords, phrases, sentences, or paragraphs from a lengthy piece of text, this approach has been explored even in Supreme Court opinions by \cite{bauer_legal_2023} or Abstractive summarisation, where the summary is composed from scratch and uses the document as a reference in order to generate a paraphrase.

Extractive summaries can be too verbose or might lack coherence due to the extraction process that is not guaranteed to be seamless. On the other hand, abstractive summaries are usually more coherent, but they are prone to hallucination which can be especially troublesome in a technical and high-stakes field like the legal one.

When dealing with this specific task, two different aspects need to be taken into account, and two separate summaries are compiled:
\begin{itemize}
    \item Facts: Using 1-2 sentences, the facts are summarised.
    \item Legal reasoning: The key arguments are highlighted from the perspective of the majority. The output is an high-level summary of what the case is about in third person. An "*" is added next to words or phrases that might be considered difficult to understand, then at the bottom of the summary, define the term.
\end{itemize}

After retrieving the distinct summaries, it is necessary to combine them and make the text readable to the majority of the public. An approach that fits this goal is called Text Style Transfer.

Style Transfer is a popular task in computer vision \cite{gatys_neural_2015} and consists in combining some content and a style taken from arbitrary images, such as creating a new painting in the style of Van Gogh. In this case, it can be used to modify one text style (Supreme Court opinions) to another, such as Twitter threads, YouTube comments, or 7th-grade-level essays.

Style Transfers into more accessible formats can improve the readability of court documents that are formatted with particular procedures and norms in mind. Despite its perks, there is a tradeoff between fidelity to the original source material and simplification. This problem arises with any sort of summarisation task but in this case in particular, it can lead to the loss of subtle nuances that legal professionals may find important.
\section{Benchmarks}
Evaluating the performance of LLMs in legal reasoning is becoming a primary topic of research, to allow a fair comparison between competitors and ensure that these models can accurately interpret statutes, analyse case law, and provide reliable legal insights.

In the next section we will examine two popular benchmarks that are being used in evaluating LLMs.
\subsection{LegalBench}
LegalBench by \cite{guha_legalbench_2023} originates from a mix of existing legal datasets (restructured for the few-shot LLM paradigm), and hand-crafted datasets created and contributed by legal professionals. The main topic regarding evaluating this kind of models is establishing what types of legal reasoning tasks can LLMs perform.

Six types of legal reasoning tasks are identified: Issue-spotting, Rule-recall, Rule-application, Rule-conclusion, Interpretation and Rhetorical-understanding.

As defined in \cite{maccormick_legal_2016}, “Legal reasoning is the process of devising, reflecting on, or giving reasons for legal acts and decisions or justifications for speculative opinions about the meaning of law and its relevance to action”. 

A common framework for executing this type of legal reasoning is the Issue, Rule, Application and Conclusion (IRAC) framework, a practical example reported from the benchmark by \cite{guha_legalbench_2023} is shown below:

\begin{enumerate}
    \item Issue Spotting: Process the issue in a given set of facts. Citing from \cite{guha_legalbench_2023}: “An issue is often either (1) a specific unanswered legal question posed by the facts, or (2) an area of law implicated in the facts.”;
    \item Rule Recall: The relevant legal rules are identified for the issue at hand. As mentioned before, the origin of such rules depends on the legal system considered;
    \item Rule Application: Consists in identifying the facts that are relevant to the rule recalled and the identified rules are applied  (rule-application);
    \item Rule Conclusion: The final step consists in determining what the legal outcome is.
\end{enumerate}

Though IRAC is one of the most famous framework for legal reasoning, another variety of legal skills might be useful, such as interpretation or rhetoric, as such the categories that the models are evaluated on are:
\begin{itemize}
    \item Issue-spotting: Tasks in which an LLM is prompted to determine whether a set of facts raises a particular set of legal questions, implicate an area of the law, or are relevant to a specific party. An example is to determine (Yes/No) whether a post on a public legal aid forum raises issues related to welfare law (i.e., public benefits or social services);
    \item Rule-recall: LegalBench has been developed following the American Common Law system, therefore recalling the correct legal rules to the respective jurisdictions is crucial. This task requires the LLM to generate the correct legal rule on an issue in a jurisdiction (e.g., the rule for hearsay in US federal court). There are two types of a rule task: open-ended generation task, in which the text of the rule for a jurisdiction must be provided, or a classification task, in which the existence of a specific rule exists in that jurisdiction must be determined. Hallucinations, as seen before, can be measured in this instance;
    \item Rule-conclusion: Testing the ability of an LLM to determine the legal outcome of a set of facts under a specified rule. For example to determine whether a contract is governed by the Uniform Commercial Code (UCC) or the common law of contracts. The relevant rule is provided via the prompt to the LLM under test;
    \item Rule-application: The LLM is prompted to provide an explanation of how the rule applies to a set of facts. Rule-application tasks were evaluated manually by a law-trained individual using a grading guide, who analysed LLM responses for both correctness and analysis: 
    \begin{itemize}
        \item Correctness: Consider, for example, an explanation which restates the rule, the fact pattern, and the predicted legal outcome. The correctness of the predicted legal outcome implies the correctness of the explanation. There is the focus on five types of errors: 
        \begin{itemize}
            \item misstatements of the legal rule
            \item misstatements of the fact pattern
            \item incorrectly asserting the legal outcome
            \item logic errors
            \item arithmetic errors
        \end{itemize}
        \item Analysis: Explanations should contain inferences from the facts that are relevant under the rule considered, and illustrate how a conclusion is reached. 
    \end{itemize}
    \item Interpretation: Interpretive tasks provide the LLM with a text, and ask the LLM to either extract a relevant piece of information, answer a question, or categorise the text by some property. An example of an interpretive task is to determine if a contractual clause contains an “audit right.”\footnote{Grants one party the right to examine and verify the other party's financial records, ensuring compliance with contractual terms (\url{https://www.contractken.com/glossary/audit-clause}};
    \item Rhetorical-understanding: An LLM is provided with a legal argument (usually excerpted from a judicial opinion), and asked to determine whether it performs a certain function or has a certain property. An example is whether a sentence from a judicial opinion provides a definition of a term.
\end{itemize}

Different metrics are used on the various types of tasks: in classification tasks the metric used is exact-match on class-balanced-accuracy. For extraction tasks, normalisation is performed on generated outputs to eliminate differences in tense, casing or punctuation. In the few tasks that requires the LLM to produce multiple classes or extracted terms per instance, F1 is used as metric, while for numerical generation tasks each test case uses one of F1 or accuracy.

In \cite{guha_legalbench_2023} the model tested were 20 both open-source and commercial and the best results were obtained by GPT-4.
\subsection{Law Bench}

As mentioned many times during this chapter, the application of these systems is heavily influenced by the country in which it is used (see \ref{fig:systems}), as such, different benchmarks for civil law systems have been developed. One notable example is LawBench by \cite{fei_lawbench_2023} constructed on the basis of the Chinese legal system.

LawBench has been crafted to assess LLMs’ legal capabilities, using as an inspiration the Bloom’s cognitive model from \cite{1f048372-501a-34de-91a6-a045447ec7fa} to classify tasks into different dimensions from three cognitive levels:
\begin{itemize}
    \item Legal knowledge memorisation: whether LLMs can memorise needed legal concepts, articles and facts, the two major types of legal knowledge that requires memorising are
    \begin{itemize}
        \item Article recitation: Given a law article number, recite the article content, including amendments;
        \item Knowledge question answering: Select the correct answer from 4 candidates given a question asking about basic legal knowledge;

    \end{itemize}
    \item Legal knowledge understanding: This batch of tests checks an LLM's ability to identify entities, events and relationships within legal text;
    \begin{itemize}
        \item Document Proofreading: Correct spelling, grammar and ordering mistakes;
        \item Dispute Focus Identification: Detect the points of dispute from the original claims and responses of the plaintiff and defendant;
        \item Marital Disputes Identification: Classification of a sentence regarding marital disputes into one of the 20 predefined dispute types;
        \item Issue Topic Identification: Classification of a user inquiry into predefined topics;
        \item Reading Comprehension: Extract relevant content from a judgement document and answer a related question; 
        \item Named-Entity Recognition: Extract a predefined set of entities (suspect, victim or evidence) from a judgement document;
        \item Opinion Summarisation: Generate a summary from a legal-related public news report;
        \item Argument Mining: Pick out of five viewpoints from the defendant, the most effective in order to dispute the given plaintiff's\footnote{A person who brings a legal action. “Plaintiff.” Merriam-Webster.com Dictionary, Merriam-Webster, \url{https://www.merriam-webster.com/dictionary/plaintiff}. Accessed 3 Mar. 2025.} perspective;
        \item Event Detection: Detect which events are mentioned in a given legal judgement document;
        \item Trigger Word Extraction: Extract which words in a given legal judgment document triggered the events reported. This task can provide a post-hoc explanation for the event identified.
    \end{itemize}

    \item Legal knowledge applying: Test the capacity of LLMs on legal reasoning based on their legal knowledge.
    \begin{itemize}
        \item Fact-based Article Prediction: Predict the article applicable based on a factual statement in a legal judgement document;
        \item Scene-based Article Prediction: Predict the article from a described scenario and a related question;
        \item Charge Prediction: Predict the charge given a fact statement;
        \item Prison Term Prediction without article: Predict the applied article number and prison term given a fact statement;
        \item Prison Term Prediction with article: Predict the prison term given the content of the applied article, charge and fact statement;
        \item Case Analysis: Select the correct answer from 4 possible responses regarding a question on a given case;
        \item Criminal Damages Calculation: Compute the amount of money involved in a case given the fact description. This task is relevant as the total amount of the crime is an important sentencing factor such as theft, financial fraud, and bribery because the sentence depends on the amount involved in the case;
        \item Consultation: Provide an appropriate answer to an user consultation;
    \end{itemize}
\end{itemize}
The metrics used to evaluate the previous tasks differ based on the type of the task, and are: 
\begin{itemize}
    \item Accuracy: Used in all single-label classification tasks (Knowledge question answering, Issue Topic Identification, Argument Mining, Case Analysis) the regression task Criminal Damages Calculation;
    \item F1: Harmonic mean of the precision and recall. This is used in all multi-label classification tasks, such as Dispute Focus Identification, Marital Disputes Identification, Event Detection, Fact-based Article Prediction, Charge Prediction;
    \item rc-F1: F1 score tailored for the Reading Comprehension task. It handles every token as a label, computes the F1 score on normalised text (removed punctuation, stories, extra whitespace and others);
    \item soft-F1: For extraction tasks Named-Entity Recognition and Trigger Word Extraction, the output might have different words with respect of the ground truth, so a soft version of F1 is used and it is computed by replacing the phrase-level exact match with the rc-F1 score, then computing the F1 on top of it.
    \item nLog-distance: For the prison term prediction tasks Prison Term Prediction with and without article, the logarithm of the difference between the extracted and ground-truth answer, then it is normalised between 0 and 1.
    \item F0.5: Used for the Document Proofreading task. The F0.5 score gives more weight to precision than to recall.
    \item Rouge-L: Used for tasks Article recitation, Opinion Summarisation, Charge Prediction and Consultation. Rouge-L considers structure similarity at sentence level and identifies longest co-occurring in sequence n-grams to compare the extracted and gold answers.
\end{itemize}

By analysing the results, as per the previous benchmark, GPT-4 and ChatGPT substantially outperform all other models and a correlation between model size and better performance in one-shot case has been found. One particular critical task is Prison Term Prediction with Article, as  the vast majority of models fail.
\section{Conclusion}

Grasping at the endless sea of the potential implications that laws have on everyone's lives and the extraordinary capabilities that Large Language Models have achieved in the last few years, I couldn’t help but wonder, if the law is all about interpretation, what happens when we hand that power over to a machine?  Many cases and publications have shown that the goal of streamlining the judicial systems is possible and Large Language Models can actually provide many benefits, but only when considering the nature of the technologies involved, acknowledging that mistakes can be made and that we should not rely blindly on the output generated, but rather consider it a suggestion. 

And so, as AI continues its cross-examination of the legal world, I had to ask myself: When it comes to justice, should we let the machines make their case, or is it time to object? 
\bibliographystyle{elsarticle-harv} 
\bibliography{bibliography}

\end{document}